\documentclass{sig-alternate-05-2015}
\usepackage{url}
\usepackage{graphicx}
\usepackage{multirow}
\usepackage[english]{babel}
\usepackage{blindtext}

\usepackage{etoolbox}
\makeatletter
\patchcmd{\maketitle}{\@copyrightspace}{}{}{}
\makeatother

\begin{document}






%

\title{Training Deep Networks for Facial Expression Recognition with Crowd-Sourced Label Distribution}
%
%
%
%
%

\numberofauthors{1}

\author{
%
%
\alignauthor
Emad Barsoum, Cha Zhang, Cristian Canton Ferrer and Zhengyou Zhang\\
       \affaddr{Microsoft Research}\\
       \affaddr{One Microsoft Way, Redmond, WA 98052}\\
       \email{\{ebarsoum, chazhang, crcanton, zhang\}@microsoft.com}
}

\maketitle
\begin{abstract}
\par Crowd sourcing has become a widely adopted scheme to collect ground truth labels. However, it is a well-known problem that these labels can be very noisy. In this paper, we demonstrate how to learn a deep convolutional neural network (DCNN) from noisy labels, using facial expression recognition as an example. More specifically, we have 10 taggers to label each input image, and compare four different approaches to utilizing the multiple labels: majority voting, multi-label learning, probabilistic label drawing, and cross-entropy loss. We show that the traditional majority voting scheme does not perform as well as the last two approaches that fully leverage the label distribution. An enhanced FER+ data set with multiple labels for each face image will also be shared with the research community. 
\end{abstract}

%
%


%
%



\section{Introduction}
\par Understanding the unspoken words from facial and body cues is a fundamental human trait, and such aptitude is vital in our daily communications and social interactions. In research communities such as human computer interaction (HCI), neuroscience and computer vision, scientists have conducted extensive research to understand human emotions. Such studies would allow creating computers that can understand human emotions as well as ourselves, and lead to seamless interactions between human and computers. 

\par Among many inputs that can be used to derive emotions, facial expression is by far the most popular. One of the pioneer works by Paul Ekman~\cite{ekman1971constants} identified 6 emotions that are universal across different cultures. Later, Ekman~\cite{ekman1977facial} developed the Facial Action Coding System (FACS), which became the standard scheme for facial expression research. Facial expression analysis can thus be conducted by analyzing facial action units for each of the facial parts (eyes, nose, mouth corners, etc.), and map them into FACS codes~\cite{tian2001recognizing}. Unfortunately, FACS coding requires professionally trained coders to annotate, and there are very few existing data sets that are available for learning FACS based facial expressions, in particular for unconstrained real-world images. 

\par With the latest advances in machine learning, it is more and more popular to recognize facial expressions directly from input images. Such appearance-based approaches have the advantage that the ground truth labels may be abundantly obtained through crowd-sourcing platforms~\cite{mturk}. The cost of tagging a holistic facial emotion is often on the order of 1-2 US cents, which is orders of magnitude cheaper than FACS coding. On the other hand, crowd-sourced labels are usually much noisier than FACS codes annotated by specially trained coders. This can be attributed to two main reasons. First, emotions are very subjective, and it is very common that two people have diametrically different opinions on the same face image. Second, the workers in crowd-sourcing platforms are paid very low, and their incentive is more on getting more work done rather than ensuring the tagging quality. Consequently, crowd-sourced labels on emotions exhibit only $65\pm5\%$ accuracy, as reported for the original FER data set~\cite{goodfellow2013challenges}. 

\par In this paper, we adopt the latest deep convolutional neural networks (DCNN) architecture, and evaluate the effectiveness of four different schemes to train emotion recognition on crowd-sourced labels. In order to overcome the noisy label issue, we asked 10 crowd taggers to re-label each image in the FER data set, resulting in a new data set named FER+\cite{ferplus}. Then, we change the cost function of the DCNN based on different schemes using the distribution of tags: majority voting, multi-label learning, probabilistic label drawing, and cross-entropy loss. We compare the performance of the trained classifiers and found the last two schemes to be the most effective to train emotion recognition classifiers based on noisy labels. 

\par The rest of the paper is organized as follows. Related works are discussed in Section~\ref{section:relatedwork} and a description of the FER+ data set is introduced in Section~\ref{section:fer}. Then, the four schemes for DCNN training are presented in Section~\ref{section:dcnn} while experimental results and conclusions are given in Section~\ref{section:exp} and \ref{section:conclusion}, respectively. 

\section{Related Work}
\label{section:relatedwork}

\par Crowd sourcing has been proven to be a cheap and effective way of tagging large amounts of data~\cite{cheap:Snow2008}. While the quality of the labels from crowd sourcing is not always guaranteed, a lot of works and investigations have been conducted to improve the tagging quality~\cite{crowdsource:Ambati2012,batliner2007, 
burmania2015}. 
For example, one effective approach is to add a gold standard ground truth as part of the dataset, i.e., data or dummy questions with known answers \cite{crowdsource:Ambati2012, eickhoffV13,kitturCS08}. Alternatively, one can filter out annotators that are too fast or too slow~\cite{CaoCKGNV14, SadoughiLB14,soleymani2010}, or use a reference set with ground truth to monitor annotators accuracy and fatigue in real-time~\cite{burmania2015}.

\par Recognizing facial expressions based on appearances has been an active research topic for decades. Early works rely on hand-crafted features such as Gabor Wavelets~\cite{zhang1999ijprai}, Local Binary Patterns on Three Orthogonal Planes (LBP-TOP) \cite{zhao2007dynamic}, Pyramid Histogram of Oriented Gradients (PHOG) \cite{ojansivu2008blur} and Local Quantized Patterns (LPQ) \cite{bosch2007representing}. Lately, due to the great success of DCNN in a wide variety of image classification tasks~\cite{alex2012, christian2015}, it has also been applied in emotion recognition \cite{kahou2013combining,liu2014facial,liu2014deeply,tang2013deep,YuZhiding2015}. One of the main attractiveness of DCNNs is its ability to learn features directly from data avoiding the tedious hand crafted feature generation used in other supervised learning methods. Hence, it is possible to have end-to-end systems that learn directly from data and infer the output with a single learning algorithm. Naturally, the quality and quantity of the training data set largely determines the overall performance of the final system. 

\par We are not the first one who realizes that the emotion of a subject is often non-exclusive. For example, in~\cite{trohidis2008multi}, Trohidis \emph{et al.} observed that music may evoke more than one emotion at the same time, and compared 4 multi-label classification algorithms to address the issue. In~\cite{devillers2005challenges}, the authors allowed the annotation of emotion mixtures for speech, and numerous works follow similar ideas in speech emotion recognition~\cite{mower2009interpreting, sobol2010classification}. In~\cite{Zhou2015:EmotionDistribution}, the authors proposed an emotion distribution learning (EDL) algorithm for still images. Their algorithm extracts LBP features from the face region, and learns a parametric model for the conditional probability distribution of emotions given an image. In contrast, our algorithm learns the features and the classifier simultaneously in a DCNN framework, thanks to a much bigger training set -- the FER+ data set. 

\begin{figure}[!t]
\centering
\includegraphics[width=\columnwidth]{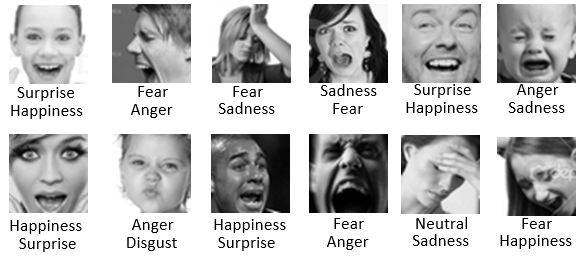}
\caption{FER vs FER+ examples. Top labels are FER and bottom labels are FER+ (after majority voting).}
\label{figure:oldvsnew}
\end{figure}

\section{The FER+ Data Set}
\label{section:fer}

\par The original FER data set was prepared by Pierre Luc Carrier and Aaron Courville by web crawling face images with emotion related keywords. The images are filtered by human labelers, but the label accuracy is not very high~\cite{goodfellow2013challenges}. A few examples are given in Figure~\ref{figure:oldvsnew}. 

\par For this paper, we decided to re-tag the FER data set with crowd sourcing. For each input image, we asked crowd taggers to label the image into one of 8 emotion types: neutral, happiness, surprise, sadness, anger, disgust, fear, and contempt. The taggers are required to choose one single emotion for each image and the gold standard method has been adopted to ensure the tagging quality. In a first attempt, tagging was stopped as long as two taggers agreed upon a single emotion, but the obtained quality was unsatisfactory. In the end, we asked 10 taggers to label each image, thus obtaining a distribution of emotions for each face image. 

\par
\begin{figure}
\centering
\includegraphics[width=0.9\columnwidth]{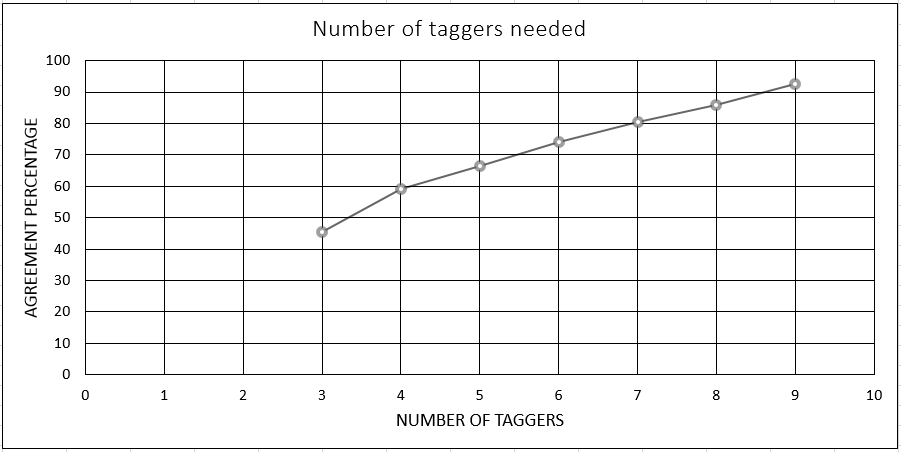}
\caption{Tagger count versus quality.}
\label{figure:TaggerCountVSQuality}
\end{figure}

 Figure~\ref{figure:TaggerCountVSQuality} shows a plot relating the tagging quality versus the number of taggers. We randomly chose 10k images in the data set and assume that the majority of the 10 labels are a good approximation to the ``ground truth'' label. Then, when we have fewer taggers, we compute how many of the majority agree with the ``ground truth'' emotion. It can be seen from the figure that when there are 3 taggers, the agreement is merely 46\%. With 5 taggers, the accuracy improves to about 67\% and, with 7 taggers, the agreement improves to above 80\%. With this, we can conclude that the number of taggers has a high impact on the final label quality~\cite{rosenthal2005conducting}. 

\par With 10 annotators for each face image, we now can generate a probability distribution of emotion capture by the facial expression, which enables us to experiment with multiple schemes during training. In section ~\ref{section:training}, we discuss in depth the 4 schemes that we tried: majority voting, multi-label learning, probabilistic label drawing, and cross-entropy loss.

\section{DCNN Learning}
\label{section:dcnn}

\par Discriminating emotion based on appearance is essentially an image classification problem. Therefore, a state-of-the-art DCNN model that performs well in image classification should also perform well in facial expression recognition. We tried multiple DCNN models, including custom versions of the VGG network~\cite{SimonyanZ14a}, GoogLeNet~\cite{Szegedy_2015_CVPR}, ResNet~\cite{he2015deep}, etc. Since comparing different DCNN models is not the objective of this paper, we adopt a custom VGG network in this paper to demonstrate emotion recognition performance on the FER+ data set. 


\begin{figure*}
\centering
\includegraphics[scale=0.35]{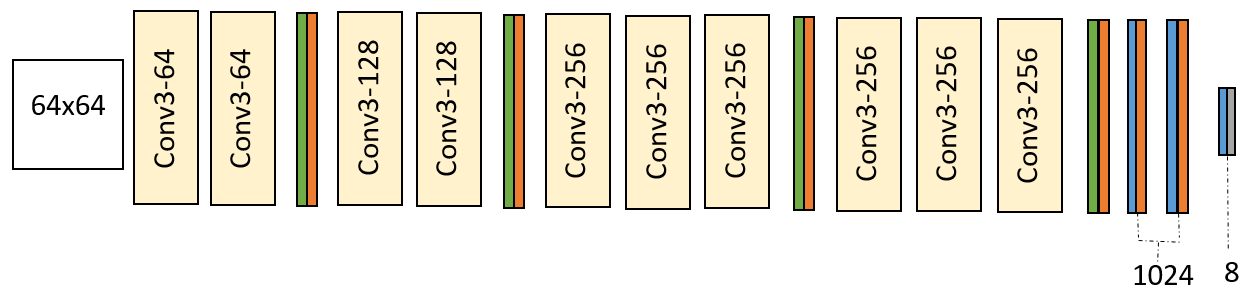}
\caption{Our custom VGG13 network: yellow, green, orange, blue and gray are convolution, max pooling, dropout, fully connected and soft-max layer, respectively.}
\label{figure:vgg13}
\end{figure*}

\subsection{Network Architecture}
\par The input to our emotion recognition model is a gray scale image at $64\times 64$ resolution. The output is 8 emotion classes: neutral, happiness, surprise, sadness, anger, disgust, fear and contempt. Our custom VGG13 model is shown in Figure~\ref{figure:vgg13}. It has 10 convolution layers, interleaved with max pooling and dropout layers. More specifically, after the input layer, there are 2 convolution layers with 64 kernels of size $3\times 3$. After max pooling, a dropout layer is added with a dropout rate of 25\%. The structure repeats but changes in the number of convolution layers and number of kernels. After all the convolution layers, 2 dense layers are added, each with 1024 hidden nodes, followed by a 50\% dropout layer. The final dense layer is followed with a soft-max layer to generate the output. 

\par Although the FER+ training set has only about 35k images, the dropout layers are effective in avoiding model overfitting in our model.

\subsection{Training}
\label{section:training}

\par We train the custom VGG13 network from scratch on the FER+ data set employing the same split between training,  validation and testing data provided in the original FER. During training we augment the data set on the fly, applying affine transforms similar to those in~\cite{YuZhiding2015}. Such data augmentation has been proven to improve the robustness of the model against translation, rotation and scaling. 

\par Thanks to the large number of taggers per image, we could generate a probability distribution for each face image. In the following, we examine how to utilize the label distribution in a DCNN learning framework during training. Let there be a total of $N$ training examples $\mathbf{I}_i,{i=1,\cdots,N}$. For the $i^{\textrm{th}}$ example, let the custom VGG13 network's output after its soft-max layer be $q_k^i$, ${k=1,\cdots,8}$, and the crowd-sourced label distribution for this example be $p_k^i$, ${k=1,\cdots,8}$. Naturally, we have: 
\begin{equation}
\sum_{k=1}^8 q_k^i = 1;\hspace{0.5cm}\sum_{k=1}^8 p_k^i = 1.
\end{equation}
We experimented 4 different schemes: majority voting (MV), multi-label learning (ML), probabilistic label drawing (PLD) and cross-entropy loss (CEL). These approaches are explained in detail below. 

\subsubsection{Majority Voting}
\par In most existing facial expression date sets, each facial image is only associated with one single emotion. It is natural to use the majority of the label distribution as the single tag for the image. More formally, we create a new target distribution $\hat{p}_k^i$ for each example $\mathbf{I}_i$, such that:  
\begin{equation}
\hat{p}_k^i =
\left\{
	\begin{array}{ll}
		1  & \mbox{if } k=\arg\max_j p_j^i \\
		0 & \mbox{otherwise}
	\end{array}.
\right.
\end{equation}
\par The cost function for DCNN learning is the standard cross-entropy cost, i.e.
\begin{equation} 
\mathcal{L} = -\sum\limits_{i=1}^N \sum\limits_{k=1}^8 \hat{p}^{i}_{k} \log q^{i}_{k}.
\end{equation}

\subsubsection{Multi-Label Learning}
\par Many face images may exhibit multiple emotions. For example, someone can be happily surprised, or angrily disgusted. The idea of multi-label learning is to admit that such multi-emotion cases exist, and it is fine for our learning algorithm to match with any of the emotions that had sufficient number of taggers labeling them. Mathematically, we adopt a new loss function as follows: 
\begin{equation} 
\mathcal{L} = -\sum\limits_{i=1}^N \arg\max_k I_\theta(p^{i}_{k}) \log q^{i}_{k},
\end{equation}
where $I_\theta(p^{i}_{k})$ is an indicator function with threshold $\theta$: 
\begin{equation}
I_\theta(p^{i}_{k}) =
\left\{
	\begin{array}{ll}
		1  & \mbox{if } p_k^i > \theta \\
		0 & \mbox{otherwise}
	\end{array}.
\right.
\end{equation}
Since more than one emotion is acceptable for each face, we let the algorithm pick the emotion it wants to train on based on the output probability of each emotion. It is basically applying multi-instance learning in the label space. Effectively, as long as the network output agrees with any of the emotions that a certain portion of the taggers agree, the cost would be low. In our experiments, the parameter $\theta$ is set to 30\%. 

\begin{table*}[h]
\centering
\vspace*{0.1cm}
\begin{tabular}{ |c|c|c|c|c|c|c| } 
  \hline
 \multirow{2}{*}{Scheme} & \multicolumn{5}{|c|}{Trials} & \multirow{2}{*}{Accuracy} \\
  \cline{2-6}
   & 1 & 2 & 3 & 4 & 5 &  \\
 \hline
 MV  & 83.60 \% & 84.89 \% & 83.15 \% & 83.39 \% & 84.23 \% &         83.852 $\pm$ 0.631 \%  \\ 
 ML  & 83.69 \% & 83.63 \% & 83.81 \% & 84.62 \% & 84.08 \% &         83.966 $\pm$ 0.362 \%  \\   
 PLD & 85.43 \% & 84.65 \% & 85.34 \% & 85.01 \% & 84.50 \% & \textbf{84.986 $\pm$ 0.366} \% \\ 
 CEL & 85.01 \% & 84.59 \% & 84.32 \% & 84.80 \% & 84.86 \% &         84.716 $\pm$ 0.239 \%  \\ 
 \hline
\end{tabular}
\caption{Testing accuracy from training VGG13 using four different schemes: majority voting (MV), multi-label learning (ML), probabilistic label drawing (PLD) and cross-entropy loss (CEL).}
\label{table:results} 
\end{table*}

\subsubsection{Probabilistic Label Drawing}
\par In the probabilistic label drawing approach, when an example is used in a training epoch, a random emotion tag is drawn from the example's label distribution $p_k^i$. We then treat this example as if it has a single emotion label as the drawn emotion tag. In the next epoch, the random drawing will happen again, and may be associated with a different emotion tag. Over the multiple epochs during training, we believe we will approach the true label distribution on average. Formally, at epoch $t$, we create a new distribution $\tilde{p}_k^i(t)$: 
\begin{equation}
\tilde{p}_k^i(t) =
\left\{
	\begin{array}{ll}
		1  & \mbox{if } k=\textrm{choice} (p_j^i) \\
		0 & \mbox{otherwise}
	\end{array},
\right.
\end{equation}
where $\textrm{choice} (p_j^i)$ is a random number generator based on the distribution $p_j^i$. The cost function for DCNN is the same standard cross-entropy loss: 
\begin{equation} 
\mathcal{L}(t) = -\sum\limits_{i=1}^N \arg\max_k \tilde{p}^{i}_{k}(t) \log q^{i}_{k}. 
\end{equation}

\subsubsection{Cross-entropy loss}
\par The fourth approach is the standard cross-entropy loss. We treat the label distribution as the target we want the DCNN to approach. That is: 
\begin{equation} 
\label{eq:crossentropy}
\mathcal{L} = -\sum\limits_{i=1}^N \sum\limits_{k=1}^8 p^{i}_{k} \log q^{i}_{k}.
\end{equation}

\section{Experimental results}
\label{section:exp}

\par We tested the above four schemes on the FER+ data set we created. As mentioned earlier, each image is tagged by 10 taggers. The label distribution is generated with a simple outlier rejection mechanism: if an emotion was tagged less than once, the frequency count of that emotion is reset to zero. The label frequencies are normalized to ensure the distribution sum to one. 

\par To compare the performance across all four approaches on the test set, we take the majority emotion as the single emotion label, and we measure prediction accuracy against the majority emotion. 

\par For each scheme, we train our custom VGG13 network 5 times, and report the accuracy numbers in Table~\ref{table:results}. Due to random initialization, the accuracy of the same scheme varies across different runs. It can be seen from the table that the PLD and CEL approaches yield the best accuracy on the test data set. Both approaches are over 1\% better in accuracy compared with MV. The t-value is around 3.1, which gives probability of 99\%-99.5\% that the statistic is significant. On the other hand, the difference between PLD and CEL is within the standard deviation. The slight advangtage of PLD may be explained by its similarity to the independently discovered DisturbLabel approach in~\cite{xie2016disturblabel}. 

\par It was a bit surprising to us that ML did not achieve as good performance as PLD and CEL. Since we ask each tagger to tag only the single dominate emotion, we expect the label distribution does not necessarily reflect the emotion distribution of the underlying image, and we thought ML would be a more flexible learning target. We hypothesize that it might be because during testing only the majority emotion is used, and there is a bigger mismatch between training and testing for ML. Further work is needed to verify our hypothesis. 

\par
\begin{figure}
\centering
\includegraphics[scale=0.20]{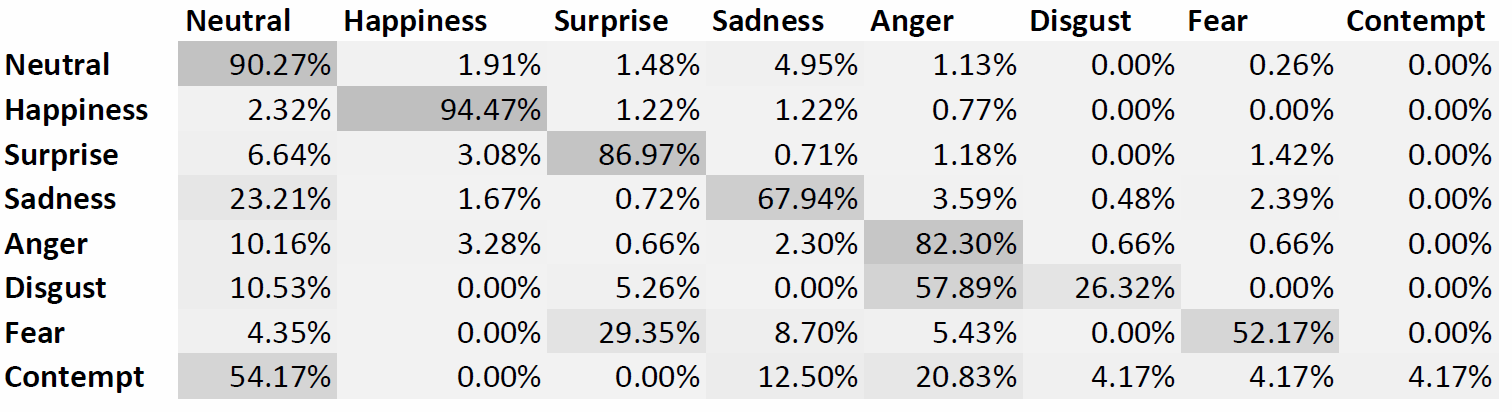}
\caption{Confusion matrix for the probability scheme.}
\label{figure:confusionmatrix}
\end{figure}

 Figure~\ref{figure:confusionmatrix} shows the confusion matrix of the best performing network. We perform well on most of the emotions except disgust and contempt. This is because we have very few examples in the FER+ training set that are labeled with these two emotions. 

\section{Conclusions}
\label{section:conclusion}

\par In this paper, we compare different schemes of training DCNN on crowd-sourced label distributions. We show that taking advantage of the multiple labels per image boost the classification accuracy compared with the traditional approach of single label from majority voting. 

The FER+ data set\cite{ferplus} is available for download in the following web address: https://github.com/Microsoft/FERPlus. 

\bibliographystyle{abbrv}
\bibliography{references} 

\end{document}